\documentclass[sigconf,nonacm]{acmart}

\AtBeginDocument{%
  \providecommand\BibTeX{{%
    \normalfont B\kern-0.5em{\scshape i\kern-0.25em b}\kern-0.8em\TeX}}}

\copyrightyear{2020}
\acmYear{2020}
\setcopyright{acmcopyright}\acmConference[MM '20]{Proceedings of the 28th ACM International Conference on Multimedia}{October 12--16, 2020}{Seattle, WA, USA}
\acmBooktitle{Proceedings of the 28th ACM International Conference on Multimedia (MM '20), October 12--16, 2020, Seattle, WA, USA}
\acmPrice{15.00}
\acmDOI{10.1145/3394171.3413727}
\acmISBN{978-1-4503-7988-5/20/10}

\usepackage{graphicx}
\usepackage{amsmath,amssymb}

\def\ie{{\it i.e.}}

\def\etc{{\it etc.}}

\def\s{{\mathbf s}}

\def\i{{\mathbf i}}
\def\x{{\mathbf x}}
\def\y{{\mathbf y}}
\def\p{{\mathbf p}}
\def\q{{\mathbf q}}

\def\P{{\mathbf P}}
\def\Pgt{{\mathbf P_\mathrm{gt}}}
\def\tP{{\widetilde{\mathbf P}}}

\def\X{{\mathbf X}}
\def\Y{{\mathbf Y}}
\def\S{{\mathbf S}}

\acmSubmissionID{1761}

\begin{document}


\title{Differentiable Manifold Reconstruction for \\ Point Cloud Denoising}

\author{Shitong Luo, Wei Hu} 
\affiliation{%
  \institution{Wangxuan Institute of Computer Technology, Peking University}
}
\email{{luost, forhuwei}@pku.edu.cn}


\thanks{Corresponding author: Wei Hu (forhuwei@pku.edu.cn). This work was supported by National Natural Science Foundation of China [61972009] and Beijing Natural Science Foundation [4194080]. }


\begin{abstract}

3D point clouds are often perturbed by noise due to the inherent limitation of acquisition equipments, which obstructs downstream tasks such as surface reconstruction, rendering and so on.  
Previous works mostly infer the displacement of noisy points from the underlying surface, which however are not designated to recover the surface explicitly and may lead to sub-optimal denoising results.  
To this end, we propose to learn the underlying manifold of a noisy point cloud from differentiably subsampled points with trivial noise perturbation and their embedded neighborhood feature, aiming to capture intrinsic structures in point clouds.
Specifically, we present an autoencoder-like neural network. 
The encoder learns both local and non-local feature representations of each point, and then samples points with low noise via an adaptive differentiable pooling operation. 
Afterwards, the decoder infers the underlying manifold by transforming each sampled point along with the embedded feature of its neighborhood to a local surface centered around the point. 
By resampling on the reconstructed manifold, we obtain a denoised point cloud. 
Further, we design an unsupervised training loss, so that our network can be trained in either an unsupervised or supervised fashion. 
Experiments show that our method significantly outperforms state-of-the-art denoising methods under both synthetic noise and real world noise.
The code and data are available at \href{https://github.com/luost26/DMRDenoise}{https://github.com/luost26/DMRDenoise}.

\end{abstract}

\ccsdesc[500]{Computing methodologies~Point-based models}
\ccsdesc[300]{Computing methodologies~3D imaging}

\keywords{point clouds, denoising, manifold, differentiable pooling}

\maketitle

\begin{figure}[h]
  \centering
  \includegraphics[width=\linewidth]{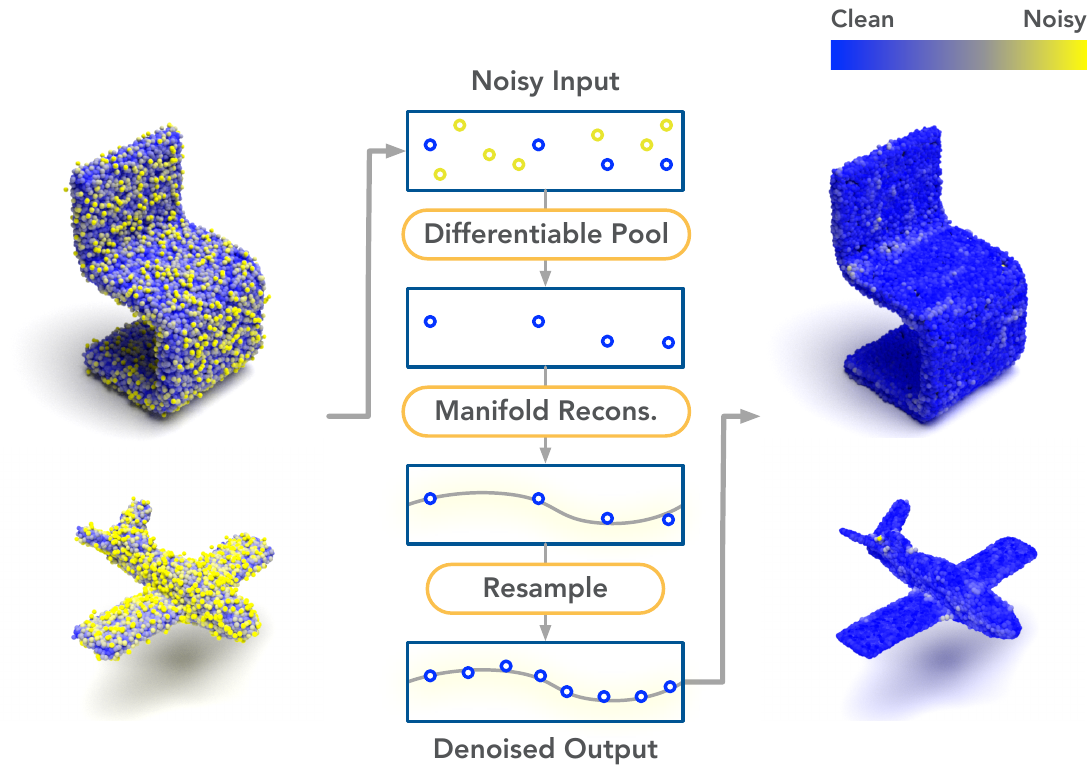}
  \caption{An overview of our method. The denoising network takes noisy point clouds as input, and then samples a subset of points with low noise via a differentiable pooling layer. Afterwards, manifolds are reconstructed based on the sampled subset of points. Finally, by sampling on the reconstructed manifold, we obtain denoised point clouds.}
  \Description{The overall architecture}
  \label{fig:teaser}
\end{figure}

\section{Introduction} 
\label{sec:introduction}

Recent advances in depth sensing, laser scanning and image processing have enabled convenient acquisition of 3D point clouds from real world scenes \footnote{Commercial products include Microsoft Kinect (2010-2014), Intel RealSense (2015-), Velodyne LiDAR (2007-2020), LiDAR scanner of Apple iPad Pro (2020), {\it etc.}}.
Point clouds consist of discrete 3D points irregularly sampled from continuous surfaces, which can be applied to a wide range of applications such as autonomous driving, robotics and immersive tele-presence. 
Nevertheless, they are often contaminated by noise due to the inherent limitations of scanning devices or matching ambiguities in the reconstruction from images, which significantly affects downstream understanding tasks since the underlying structures are deformed. 
Hence, point cloud denoising is crucial to relevant 3D vision applications, which is also challenging due to the irregular and unordered characteristics of point clouds. 

Previous point cloud denoising methods include non-deep-learning based methods \cite{bilat, jetsfit2005, MRPCA2017, GLR2019, projbased2018, featuregraph2020} and deep-learning based methods \cite{NeuralProj2019, PCN2020, TotalDenoising2019, ECNet2018}.  
We focus on the class of deep-learning based methods, which have achieved promising denoising results thanks to the advent of neural network architectures crafted for point clouds \cite{qi2017pointnet,qi2017pointnet2,yang2018foldingnet,achlioptas2018learning,ECNet2018, te2018rgcnn,PPPU2019, yu2018pu, wang2019dynamic, gao2020graphter}. 
Neural Projection \cite{NeuralProj2019}, PointCleanNet \cite{PCN2020} and Total Denoising \cite{TotalDenoising2019} are pioneers of deep-learning based point cloud denoising approaches. 
In general, these methods infer the displacement of noisy points from the underlying surface and reconstruct {\it points}, which however are not designated to recover the surface explicitly and may lead to sub-optimal denoising results.

To this end, inspired by that a point cloud is typically a representation of some underlying surface or 2D manifold over a set of sampled points, we propose to explicitly learn the underlying {\it manifold} of a noisy point cloud for reconstruction, aiming to capture intrinsic structures in point clouds. 
As demonstrated in Fig.~\ref{fig:teaser}, the key idea is to sample a subset of points with low noise (\ie, closer to the clean surface) via differentiable pooling, and then reconstruct the underlying manifold from these points and their embedded neighborhood features.
By resampling on the reconstructed manifold, we obtain a denoised point cloud.

In particular, we present an autoencoder-like neural network for differentiable manifold reconstruction. 
At the encoder, we learn both local and non-local features of each point, which embed the representations of local surfaces. 
Based on the learned features, we sample points that are closer to the underlying surfaces (less noise perturbation) via the proposed adaptive {\it differentiable pooling} operation, which narrows down the latent space for reconstructing the underlying manifold.  
These sampled points are pre-filtered and retained, while the other points are discarded. 
At the decoder, we infer the underlying manifold by transforming each sampled point along with the embedded neighborhood feature to a local surface centered around the point---referred to as {\it "patch manifold"}. 
By sampling on such patch manifolds, we finally obtain a denoised point set which captures intrinsic structures of the underlying surface. 
Further, we design an unsupervised training loss, so that our network can be trained in either an unsupervised or supervised fashion. 
Experiments show that our method significantly outperforms state-of-the-art denoising methods especially at high noise levels.

To summarize, the contributions of our paper include
\begin{itemize}
    \item We propose a differentiable manifold reconstruction paradigm for point cloud denoising, aiming to learn the underlying manifold of a noisy point cloud via an autoencoder-like framework. 
    
    \item We propose an adaptive differentiable pooling operator on point clouds, which samples points that are closer to the underlying surfaces and thus narrows down the latent space for reconstructing the underlying manifold. 
    
    \item We infer the underlying manifold by transforming each sampled point along with the embedded feature of its neighborhood to a local surface centered around the point---a patch manifold.

    \item We design an unsupervised training loss, so that our network can be trained in either an unsupervised or supervised fashion.
\end{itemize}

\section{Related Work}


\subsection{Non-deep-learning Based Point Cloud Denoising}
\label{sec:related-nondl}
Non-deep-learning based point cloud denoising methods have been extensively studied, which mainly include local-surface-fitting based methods, sparsity based methods and graph based methods.

\begin{itemize}
    \item \textbf{Local-surface-fitting based methods.} This class of methods approximate the point cloud with a smooth surface and then project points in the noisy point cloud onto the fitted surface. \cite{MLS2001} proposes a moving least squares (MLS) projection operator to calculate the optimal fitting surface of the point cloud. Similarly, other surface fitting methods have been proposed for point cloud denoising such as jet-fitting with re-projection \cite{jetsfit2005} and bilateral filtering \cite{bilat} which take into account both point coordinates and normals. 
    However, these methods are often sensitive to outliers.
    
    \item \textbf{Sparsity based methods.} This class of methods are based on the sparse representation theory \cite{xu2015sparsity, sparsecoding2010, sun2015denoising}. They generally reconstruct normal vectors by solving an optimization problem of sparse regularization and then update the position of points based on the reconstructed normals. Moving Robust Principal Component Analysis (MRPCA) \cite{MRPCA2017} is a recently proposed sparsity-based method. However, the performance tends to degrade when the noise level is high due to over-smoothing or over-sharpening.
    
    \item \textbf{Graph based methods.} This class of methods represent point clouds on graphs and perform denoising via graph filters \cite{graphbased2015, graphbased2018, GLR2019, featuregraph2020,Hu2020gsp}. In \cite{graphbased2015}, the input point cloud is represented as signal on a $k$-nearest-neighbor graph and then denoised via a convex optimization problem regularized by the gradient of the graph. In \cite{GLR2019}, Graph Laplacian Regularization (GLR) of low dimensional manifold models is proposed for point cloud denoising.
\end{itemize}


\subsection{Deep-learning Based Point Cloud Denoising}

With the advent of point-based neural networks \cite{qi2017pointnet, qi2017pointnet2, wang2019dynamic}, deep point cloud denoising has received increasing attention. Existing deep learning based methods generally involve predicting the displacement of each point in noisy point clouds via neural networks, and apply the inverse displacement to each point. 

Among them, Neural Projection \cite{NeuralProj2019} employs PointNet \cite{qi2017pointnet} to predict the tangent plane at each point, and projects points to the tangent planes. However, training a Neural Projection denoiser requires the access to not only clean point clouds but also normal vectors of each point. 

PointCleanNet \cite{PCN2020} predicts displacement of points from the clean surface via PCPNet \cite{guerrero2018pcpnet}---a variant of PointNet. It is trained end-to-end by minimizing the $\ell 2$ distance between the denoised point cloud and the ground truth, which does not require the access to normal vectors. PointCleanNet out-performs some classical denoising methods including bilateral filtering and jet fitting. The main defect of PointCleanNet includes outlier sensitivity and point cloud shrinking.

Total Denoising \cite{TotalDenoising2019} is the first unsupervised deep learning method for point cloud denoising. It is based on the assumption that points with denser surroundings are closer to the underlying surface. Hence, it introduces a spatial prior that steers convergence towards the underlying surface without the supervision of ground truth point clouds. However, the unsupervised denoiser is sensitive to outliers and may shrink point clouds. 

In addition to denoising networks, some other neural network architectures involve point cloud consolidation, which includes denoising but is often only applicable to trivial noise. PointProNet \cite{PointProNets2018} projects patches in the point cloud into 2D height maps and leverages a 2D CNN to denoise and upsample them. EC-Net \cite{ECNet2018} and 3PU \cite{PPPU2019} mainly focus on upsampling, and have shown to be robust against trivial noise. These consolidation methods are generally prone to fail when the noise level is high \cite{PCN2020}.



\section{Method}
In this section, we present our method on learning the underlying manifold for point cloud denoising. 
We start with an overview of our key ideas, and then elaborate on the proposed differentiable manifold reconstruction. 
Finally, we present our loss functions as well as provide further analysis into our method.


\subsection{Overview}
\begin{figure*}
  \centering
  \includegraphics[width=\linewidth]{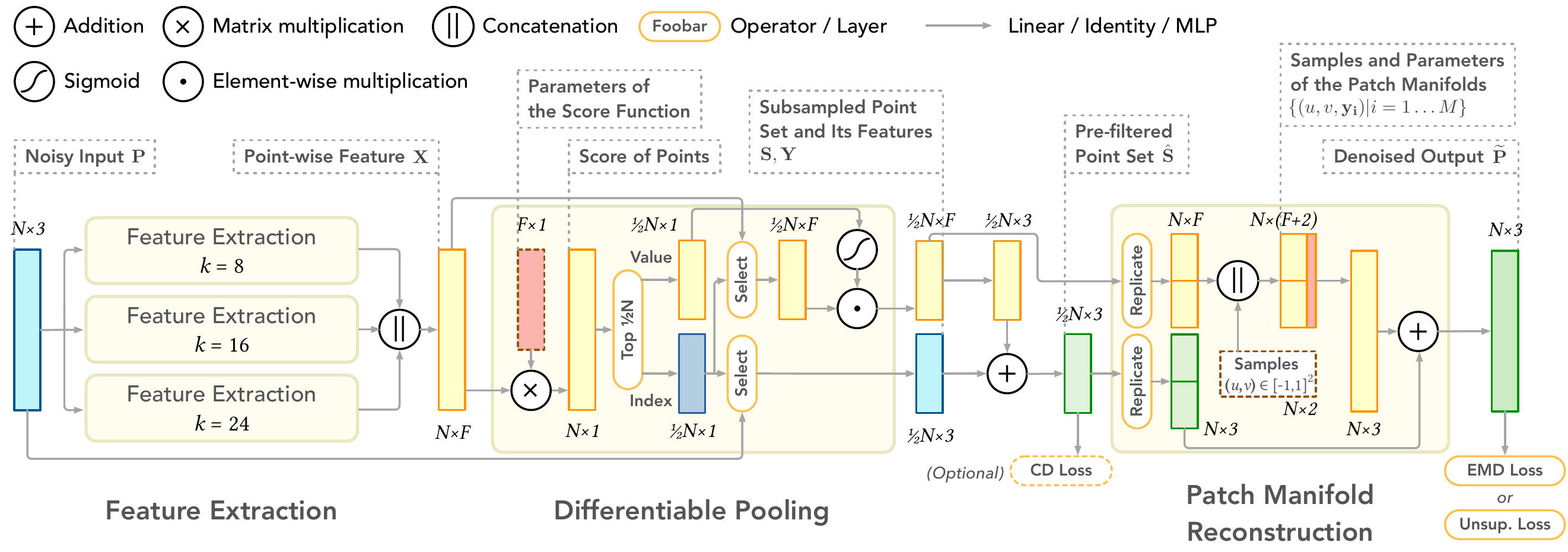}
  \caption{Illustration of the proposed point cloud denoising framework.}
  \Description{The neural network.}
  \label{fig:framework}
\end{figure*}

Given an input point cloud $\P \in \mathbb{R}^{N \times 3} $ corrupted by noise, our network produces a clean point cloud $\tP \in \mathbb{R}^{N \times 3} $.
As illustrated in Fig.~\ref{fig:framework}, we propose an autoencoder-like network architecture for denoising.

\begin{itemize}
    \item \textbf{Representation Encoder $\mathcal E$.}  $\mathcal E$ samples a subset of $M$ points $\S \in \mathbb{R}^{M \times 3}$ that are perturbed by less noise from $\P$ via differentiable pooling. 
    Specifically, $\mathcal E$ consists of a feature extraction unit and a differentiable downsampling (pooling) unit. 
    The feature extraction unit produces features that encode both local and non-local geometry at each point of $\P$. The extracted features are then fed into the differentiable pooling operator---essentially a downsampling unit that identifies points that are closer to the underlying surface, leading to a subset of points $\S$.
    
    \item \textbf{Manifold Reconstruction Decoder $\mathcal D$.} $\mathcal D$ first infers the underlying manifold from $\S$ and then samples on the inferred manifold to produce the denoised point set $\tP \in \mathbb{R}^{N \times 3} $. 
    We transform each point in $\S$ along with the embedded neighborhood feature to a local surface centered around each point---a {\it patch manifold}. 
    By sampling multiple times on each patch manifold, we reconstruct a clean point cloud $\tP$. 
\end{itemize}

Further, we propose a dual supervised loss function as well as an unsupervised loss, so that our network can be trained end-to-end in an unsupervised or supervised fashion.

\subsection{Representation Encoder with Differentiable Pooling}
\label{sec:encoder}
The representation encoder consists of a feature extraction unit and a differentiable pooling unit, which we discuss in details as follows.

\subsubsection{Feature Extraction Unit} 
The feature extraction unit consists of multiple dynamic graph convolution layers, leveraging on the DGCNN \cite{wang2019dynamic}. 
Given features $\X^{\ell}=\{\x_i^{\ell}\}_{i=1}^N \in \mathbb{R}^{N \times F^\ell}$ in the $\ell$th layer, the $(\ell+1)$th layer first constructs a $k$-Nearest-Neighbor ($k$-NN) graph based on the Euclidean distance between features, and then performs edge convolution \cite{wang2019dynamic} on the graph:
\begin{equation} \label{eq:edgeconv}
\x_i^{\ell+1} = G_{\ell}(\X^{\ell}) = \operatorname{ReLU}\big( \max_{j \in \mathcal{N}(i)} H_\theta(\x_i^{\ell}, \x_j^{\ell} - \x_i^{\ell}) \big).
\end{equation}
Here, $H_\theta$ is a densely connected multi-layer perceptron (MLP) parameterized by $\theta$, $\mathcal{N}(i)$ denotes the neighborhood of point $i$, and $\max$ is the element-wise max pooling function. 

To capture higher-order dependencies, multiple dynamic graph convolution layers are chained and connected densely within a feature extraction unit \cite{PPPU2019, liu2019densepoint, huang2017densely, li2019deepgcns}:
\begin{equation} \label{eq:graphconv}
\X^{\ell} = G_{\ell} \big([ \X^{\ell-1}, \ldots, \X^1, \X^0 ]\big),
\end{equation}
where $[\ldots]$ denotes the concatenation operation, and $\X^0$ is the input feature.

As depicted above, we adopt dense connections both within and between graph convolution layers. Within graph convolution layers, the MLP $H_\theta$ is densely connected --- each fully connected (FC) layer's output is passed to all subsequent FC layers. Between graph convolution layers, the features output by each layer are fed to all subsequent layers. Dense connections reduce the number of the network's parameters and produce features with richer contextual information \cite{PPPU2019,liu2019densepoint}.

In addition, we assemble multiple feature extraction units with different $k$-NN values in parallel to obtain features of different scales, and concatenate them before passing them to downstream networks. 
The final output is a feature matrix $\X \in \mathbb{R}^{N \times F}$, where $N$ denotes the number of points and $F$ denotes the dimension of features.


\subsubsection{Differentiable Pooling Operator}

Having extracted multi-scale features from the input point cloud $\P$, we propose a differentiable pooling operator on point clouds to sample a subset of points $\S$ from $\P$ adaptively. 
Ideally, the operator will learn to identify points that are closer to the underlying surface, which capture the surface structure better and thus will be used to reconstruct the underlying manifold at the decoder.
Different from existing pooling techniques that often employ hand-crafted rules such as random sampling and farthest point sampling \cite{qi2017pointnet2}, our differentiable pooling learns the optimal downsampling strategy adaptively during the training process.

Now we formulate the differentiable pooling operator. 
Given the learned feature $\X \in \mathbb{R}^{N \times F}$ of the input point cloud $\P$ obtained from the feature extraction unit, our pooling operator first computes a score for each point:
\begin{equation} \label{eq:score}
\s = \operatorname{Score}(\X),
\end{equation}
where $\operatorname{Score}(\cdot)$ is the score function implemented by an MLP that produces a score vector $\s \in \mathbb{R}^{N\times 1}$. 
The score function will learn a higher score for points closer to the underlying surface and a lower score for points perturbed with large noise during the end-to-end training process.

Points in $\P$ that have top-$M$ ($M<N$) scores will be retained, while the others will be discarded:
\begin{equation} \label{eq:gpool10}
\i = \operatorname{arg \ top}_M (\s),
\end{equation}
\begin{equation} \label{eq:gpool1}
\S = \P[\i],
\end{equation}
where $\i$ is the index vector of the top-$M$ points and $\S \in \mathbb{R}^{M\times 3}$ is the downsampled point set. 
In the experiments, we set $M=\frac{N}{2}$ without loss of generality. 

To make the score function differentiable so as to be trained by back propagation \cite{gao2019gpool}, we deploy the following gate operation on the features of the sampled point set $\X[\i]$ to acquire the features $\Y$ of $\S$:
\begin{equation} \label{eq:gpool2}
    \Y = \X[\i] \odot \operatorname{sigmoid}(\s[\i] \cdot \mathbf{1}^{1\times F}),
\end{equation}
where $\Y \in \mathbb{R}^{M \times F}$ is the feature matrix of $\S$ after the above gate operation, $\X[\i] \in \mathbb{R}^{M \times F} $ is the feature matrix of $\S$ before the gate operation, $\s[\i] \in \mathbb{R}^{M \times 1}$ is the score vector of the retained points, and $\odot$ denotes element-wise multiplication.

To further reduce the noise variance of the sampled point set $\S$, we perform pre-filtering on $\S$:
\begin{equation} \label{eq:gpool3}
    \hat{\S} = \S + \Delta \S , \\
\end{equation}
\begin{equation}
    \Delta \S = \operatorname{MLP}(\Y),
\end{equation}
where $\hat{\S}, \Delta \S \in \mathbb{R}^{M\times 3}$, and $\Delta \S$ is produced by an MLP that takes the feature matrix $\Y$ as input. The pre-filtering term $\Delta \S$ moves each point in $\S$ closer to the underlying surface, which will lead to more accurate manifold reconstruction at the decoder to be discussed.


\subsection{Manifold Reconstruction Decoder}

\begin{figure}
  \centering
  \includegraphics[width=\linewidth]{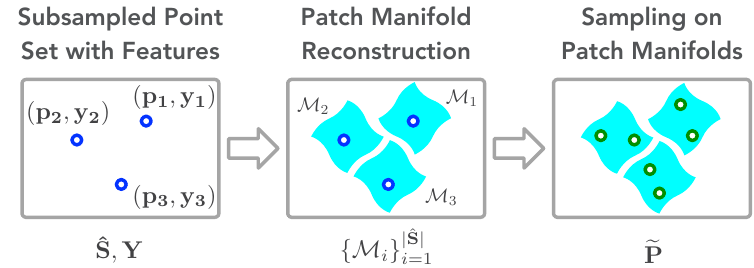}
  \caption{Illustration of the patch manifold reconstruction and resampling. Note that $\tP$ is resampled from the manifolds, so there is no strict point-to-point correspondence between $\hat{\S}$ and $\tP$.}
  \Description{Patch manifold reconstruction.}
  \label{fig:patch_m}
\end{figure}

The manifold reconstruction decoder transforms each point in the pre-filtered low-noise point set $\hat{\S}$ along with its embedded neighborhood feature matrix $\Y$ into a local surface centered around the point---referred to as a {\it patch manifold}.  
Afterwards, we upsample $\hat{\S}$ to a denoised point cloud $\tP \in \mathbb{R}^{N \times 3}$ based on the inferred patch manifolds. The whole process is illustrated in Fig. ~\ref{fig:patch_m}.

As discussed in Sec.~\ref{sec:encoder}, a feature vector $\y_i \in \mathbb{R}^F$ encodes the geometry of the {\it neighborhood surface} surrounding the point $\p_i \in \hat{\S}$, so that $\y_i$ can be transformed into a manifold that describes the local underlying surface around $\p_i$. 
We refer to such locally defined manifold as a {\it patch manifold} around $\p_i$. 

Formally, we first define a 2D manifold $\mathcal{M}$ embedded in the 3D space parameterized by some feature vector $\y$ as:
\begin{equation}
\mathcal{M}(u, v; \y) : [-1,1] \times [-1, 1] \rightarrow \mathbb{R}^3, 
\label{eq:p_manifold}
\end{equation}
where $(u,v)$ is some point in the 2D rectangular area $[-1, 1]^2$.
Eq.~\eqref{eq:p_manifold} maps the 2D rectangle to an arbitrarily shaped patch manifold parameterized by $\y$. 
Such mapping allows us to draw samples from the arbitrarily shaped patch manifold $\mathcal{M}$ in the following way: 
we firstly draw samples from the uniform distribution over $[-1, 1]^2$ and then transform them into the 3D space via the mapping.

Having defined a mapping to manifold $\mathcal{M}$, it is natural to define the patch manifold $\mathcal{M}_i$ around each point $\p_i$ in $\hat{\S}$ as:
\begin{equation}
\mathcal{M}_i(u,v;\y_i) = \p_i + \mathcal{M}(u, v; \y_i),
\end{equation}
which moves the constructed manifold $\mathcal{M}(u, v; \y_i)$ to a local surface centering at $\p_i$.    

Now we have a set of patch manifolds $\{\mathcal{M}_i | \p_i \in \hat{\S}\}_{i=1}^M$, which characterize the underlying surface of the point cloud. 
By sampling on these M patch manifolds, we can obtain the denoised point set $\tP$.

Specifically, we assume the number of points in the subsampled point set is the half of that in the input point set, \ie, {M=$|\hat{\S}| = \frac{1}{2}|\P|$}. In order to acquire a denoised point set $\tP$ that has the same size as the input point set $\P$, we need to sample twice on each patch manifold. Hence, it is essentially an upsampling process.

In practice, the parameterized patch manifold $\mathcal{M}_i(u, v; \y_i)$ is implemented by an MLP:
\begin{equation}
\mathcal{M}_i(u,v;\y_i) = \operatorname{MLP}_{\mathcal{M}}([u, v, \y_{i}]).
\end{equation}
We choose the MLP implementation because it is a universal function approximator \cite{leshno1993multilayer} which is expressive enough to approximate arbitrarily shaped manifolds.

Then, we sample two points from each patch manifold $\mathcal{M}_i([u, v, \y_i])$, leading to a denoised point cloud :
\begin{equation}
\tP = \begin{pmatrix}
p_1 + \operatorname{MLP}_\mathcal{M}([u_{11}, v_{11}, \y_1]) \\
p_1 + \operatorname{MLP}_\mathcal{M}([u_{12}, v_{12}, \y_1]) \\
\vdots \\
p_M + \operatorname{MLP}_\mathcal{M}([u_{M1}, v_{M1}, \y_M]) \\
p_M + \operatorname{MLP}_\mathcal{M}([u_{M2}, v_{M2}, \y_M])
\end{pmatrix}.
\end{equation}

To summarize, by learning a parameterized patch manifold $\mathcal{M}(u,v;\y_i)$, $ i = 1, \ldots ,M = | \hat{\S}|$ from each point $i$ in $\hat{\S}$ and sampling on each patch manifold, we reconstruct a clean point cloud from the noisy input.


\subsection{Loss Functions}
We present loss functions for supervised training and unsupervised training respectively.

\subsubsection{Supervised Training Loss} 
We consider {\it dual loss} in the setting of supervised training to measure the quality of both subsampling and final point cloud reconstruction. That is, we have two parts in the supervised loss function, including
1) a loss function $\mathcal{L}_{\mathrm{sample}}$ to quantify the distance between the subsampled and pre-filtered set $\hat{\S}$ and the ground truth point cloud $\P_\mathrm{gt}$, which explicitly reduces the noise in $\hat{\S}$ but is not required for the convergence of the training; 
2) a loss function $\mathcal{L}_{\mathrm{rec}}$ to quantify the distance between the finally reconstructed point cloud $\tP$ and the ground truth $\P_\mathrm{gt}$. 

Formally, our network can be trained supervisedly end-to-end by minimizing 
\begin{equation}
    \min_{\Theta} \mathcal{L}_{\mathrm{sample}} + \mathcal{L}_{\mathrm{rec}},
\end{equation}
where $\Theta$ denotes the learnable parameters in the network.

We choose the Chamfer distance (CD) \cite{fan2017pointsetgen} as $\mathcal{L}_{\mathrm{sample}}$, since $\hat{\S}$ and $\P_\mathrm{gt}$ exhibit different number of points, \ie, $|\hat{\S}| < |\P_\mathrm{gt}|$. It is defined as
\begin{equation}
\label{eq:cd_train}
\mathcal{L}_{\mathrm{sample}} = \mathcal{L}_{\mathrm{CD}}(\hat{\S},\P_\mathrm{gt})= \frac{1}{\left|\hat{\S}\right|} \sum_{\p \in \hat{\S}} \min _{\q \in \P_\mathrm{gt}}\| \p - \q\|_2^2+\frac{1}{\left|\P_\mathrm{gt}\right|} \sum_{\q \in \P_\mathrm{gt}} \min _{\p \in \hat{\S}}\| \q - \p\|_2^2.
\end{equation}
This loss term improves the denoising quality by explicitly optimizing the sampled and pre-filtered set $\hat{\S}$, but is optional for the network training. 



We choose the Earth Mover's distance (EMD) \cite{fan2017pointsetgen} as $\mathcal{L}_{\mathrm{rec}}$, which is shown superior to the Chamfer distance in terms of the visual quality \cite{liu2019morphing, achlioptas2018learning}. 
The Earth Mover's distance is defined when two point clouds have the {\it same} number of points. 
Fortunately, the denoising task naturally satisfies this requirement. 
The EMD loss measuring the distance between the denoised point cloud $\tP$ and the ground truth point cloud $\P_\mathrm{gt}$ is given by:
\begin{equation}
\mathcal{L}_{\mathrm{rec}}=\mathcal{L}_{\mathrm{EMD}}\left(\tP, \P_\mathrm{gt}\right)=\min _{\varphi: \tP \rightarrow \P_\mathrm{gt}} \frac{1}{N} \sum_{\p \in \tP}\|\p-\varphi(\p)\|_{2}^2,
\end{equation}
where $N = |\tP| = |\P_\mathrm{gt}|$, and $\varphi$ is a bijection.

Note that, previous works on denoising \cite{PCN2020, TotalDenoising2019} often suffer from the clustering effect of points, which is alleviated by introducing a repulsion loss. 
Our architecture does not suffer from this problem thanks to the one-to-one correspondence of points in $\mathcal{L}_\mathrm{EMD}$.

\subsubsection{Unsupervised Training Loss}
Our network can also be trained in an unsupervised fashion. 
Leveraging the unsupervised denoising loss in \cite{TotalDenoising2019}, we design an unsupervised loss tailored for our manifold reconstruction based denoising.  
The key observation is that points with a denser neighborhood are closer to the underlying clean surface, which may be regarded as ground truth points for training a denoiser. 

In \cite{TotalDenoising2019}, the unsupervised denoising loss is defined as 
\begin{equation}
\label{eq:unsup}
\mathcal{L}_{\mathrm{U}} = \frac{1}{N} \sum_{i=1}^{N} \mathbb{E}_{\q \sim P(\q | \p_i)} \| f(\p_i) - \q \|,
\end{equation}
where $P(\q|\p_i)$ is a prior capturing the probability that a point $\q$ from the noisy point cloud is the underlying clean point of the given $\p_i$ in the noisy point cloud.
Empirically, $P(\q|\p_i)$ is defined as $ P(\q|\p_i) \propto \exp ( - \frac{\| \q - \p_i \|_2^2}{2\sigma^2} )$, so that sampling points from the noisy input point cloud $\P$ according to $P(\q|\p_i)$ produces points that are closer to the underlying clean surface with high probability \cite{TotalDenoising2019}.
$f(\cdot)$ represents the denoiser that maps a noisy point $\p_i$ to a denoised point $\q_i$. It is a bijection between the noisy point cloud $\P$ and the output point cloud $\tP$. 

This bijection can be naturally established in previous deep-learning based denoising methods such as PointCleanNet \cite{PCN2020}, Total Denoising \cite{TotalDenoising2019} \etc, because these methods predict the displacement of each point. 
However, there is no natural one-to-one correspondence between $\P$ and $\tP$ in our method based on patch manifolds. 
Hence, we seek to construct one by
\begin{equation}
f=\arg\min _{f: \P \rightarrow \tP} \sum_{\p \in \P}\|f(\p) - \p\|_{2}.
\end{equation}
Having established the bijection $f$ between $\P$ and $\tP$, the unsupervised loss $\mathcal{L}_{\mathrm{U}}$ in Eq.~\eqref{eq:unsup} can be computed.



\subsection{Analysis}
Intuitively, our method can be regarded as a generalization of local-surface-fitting based denoising methods. 
As discussed in Section~\ref{sec:related-nondl}, local-surface-fitting based methods divide point cloud into patches and fit each patch via approximations.
The {\it patch manifold} defined in our method is essentially a local surface surrounding some point in the subsampled point set $\hat{\S}$,  which is analogous to patches in local-surface-fitting based methods. Our manifold reconstruction decoder leverages on neural networks to infer the shape of patch manifolds, which is analogous to patch fitting.

Another intuitive interpretation of our method is that, our differentiable pooling layer is analogous to a low-pass filter which removes high-frequency components (\ie, noise), while the manifold reconstruction is similar to high-pass filtering which recovers details from the embedded neighborhood features to avoid over-smoothing.

\begin{table*}
  \caption{Comparison of denoising algorithms. Each resolution and noise level is evaluated by 60 point clouds of different shapes from our collected test dataset, which is a subset of ModelNet-40. }
  \begin{tabular}{l|cccccccc|cccccccc}
\toprule
\# Points & 
\multicolumn{8}{c|}{20K} & 
\multicolumn{8}{c}{50K} \\
Noise & 
\multicolumn{2}{c}{1\%} & \multicolumn{2}{c}{2\%} & 
\multicolumn{2}{c}{2.5\%} & \multicolumn{2}{c|}{3\%} &

\multicolumn{2}{c}{1\%} & \multicolumn{2}{c}{2\%} & 
\multicolumn{2}{c}{2.5\%} &\multicolumn{2}{c}{3\%} 

\\
$10^{-2}$ & 
CD & P2S & CD & P2S & CD & P2S & CD & P2S &
CD & P2S & CD & P2S & CD & P2S & CD & P2S \\
\midrule
Bilateral \cite{bilat} & 
1.54 & 1.27 & 1.84 & 1.82 & 2.11 & 2.26 & 2.43 & 2.78 &
1.04 & 0.94 & 1.61 & 1.90 & 1.97 & 2.49 & 2.37 & 3.17  \\
Jet \cite{jetsfit2005} & 
1.25 & 0.96 & 2.11 & 2.32 & 2.55 & 3.04 & 2.99 & 3.78 &
1.11 & 1.10 & 2.01 & 2.61 & 2.44 & 3.35 & 2.86 & 4.09 \\
MRPCA \cite{MRPCA2017} & 
\textbf{1.13} & \textbf{0.72} & 2.12 & 2.18 & 2.66 & 3.02 & 3.16 & 3.84 &
1.03 & 0.91 & 2.12 & 2.63 & 2.58 & 3.42 & 3.02 & 4.18 \\
GLR \cite{GLR2019} & 
1.16 & 0.88 & 1.78 & 1.87 & 2.20 & 2.55 & 2.65 & 3.30 &
0.94 & 0.88 & 1.79 & 2.28 & 2.24 & 3.05 & 2.68 & 3.83 \\
\midrule
TotalDn \cite{TotalDenoising2019} & 
1.51 & 1.23 & 2.57 & 2.97 & 3.02 & 3.75 & 3.46 & 4.51 &
1.13 & 1.03 & 2.20 & 2.80 & 2.66 & 3.60 & 3.09 & 4.37 \\
PCNet \cite{PCN2020} & 
1.45 & 1.20 & 2.25 & 2.41 & 2.79 & 3.23 & 3.37 & 4.12 &
0.95 & \textbf{0.74} & 1.41 & 1.37 & 2.03 & 2.19 & 2.86 & 3.28 \\
\midrule
Ours (Supervised) & 
\textbf{1.14} & \textbf{0.85} & 
\textbf{1.40} & \textbf{1.16} & 
\textbf{1.50} & \textbf{1.37} & 
\textbf{1.79} & \textbf{1.67} &

\textbf{0.84} & \textbf{0.74} & 
\textbf{1.09} & \textbf{1.11} & 
\textbf{1.39} & \textbf{1.48} & 
\textbf{1.92} & \textbf{2.32} \\

Ours (Unsupervised) & 
1.45 & 1.35 & 1.82 & 1.92 & 2.07 & 2.22 & 2.32 & 2.71 &
1.14 & 1.23 & 1.65 & 2.14 & 1.89 & 2.59 & 2.22 & 3.21  \\
\bottomrule
\end{tabular}

  \label{tab:comp}
\end{table*}

\section{Experimental Results} \label{sec:experiments}

\begin{table}
  \caption{Comparison of different denoising methods on the point clouds generated by simulated LiDAR scanning with realistic LiDAR noise. LiDAR noise is an unseen noise pattern to our denoiser since we train our denoiser only on Gaussian noise. Results show that our denoiser is \textit{\textbf{effective in generalizing to unseen noise pattern}}, and its generalizability is better than other denoisers.}
  \begin{tabular}{c|cccc|cc|c}
\toprule
$10^{-2}$ & Bilat. & Jet & MRPCA & GLR & TotalDn & PCN  & Ours \\
\midrule
CD & 
1.23 & 1.18 & 1.10 & \textbf{1.06} &
1.25 & 1.09 &
\textbf{1.06} \\
P2S & 
1.13 & 1.17 & 0.99 & 0.99 &
1.17 & 0.93 &
\textbf{0.88}\\
\bottomrule
\end{tabular}

  \label{tab:blensor}
\end{table}



In this section, we compare our method quantitatively and qualitatively with state-of-the-art denoising methods.

\subsection{Experimental Setup}

\noindent \textbf{Dataset.} For training, we have collected 13 different classes with 7 different meshes, each from ModelNet-40 \cite{wu2015modelnet}. 
We use Poisson disk sampling to sample points from the meshes, at resolution levels ranging from 10K to 50K points. The point clouds are then perturbed by Gaussian noise with standard deviation from 1\% to 3\% of the bounding box diagonal. Due to the limit of GPU memory, we split the point clouds into patches consisting of 1024 points and feed them into the neural network.

For testing, we have collected 20 classes with 3 meshes each, which are different from the training set. Similarly, we use Poisson disk sampling at resolution levels of 20K and 50K points to generate point clouds and perturb them by Gaussian noise with standard deviation of  1\%, 2\%, 2.5\% and 3\% of the bounding box diagonal, leading to 8 classes, each with 60 point clouds. 

Furthermore, to examine the generalizability of our method to unseen noise patterns, we also generate 60 noisy point clouds via LiDAR simulators. The simulator we use is the simulation package Blensor \cite{gschwandtner2011blensor}, which can produce more realistic point clouds and noise. We use Velodyne HDL-64E2 as the scanner model in simulations. Similar to training, we split the point clouds into patches using the K-means algorithm, and feed them separately into the denoiser.

For qualitative evaluation, we additionally use the {\it Paris-rue-Madame} dataset \cite{serna2014paris}, which is obtained from the real world via laser scanners.
\newline

\noindent \textbf{Metrics.} We use the Chamfer distance (CD) \cite{fan2017pointsetgen} between the ground truth point cloud $\Pgt$ and the output point cloud $\tP$ as an evaluation metric:
\begin{equation}
\label{eq:cd_eval}
\mathcal{C}(\Pgt,\tP) = \frac{1}{\left| \tP \right|} \sum_{\q \in \tP} \min _{\p \in \Pgt}\|\q - \p\|_2 + \frac{1}{\left| \Pgt \right|} \sum_{\p \in \Pgt} \min _{\q \in \tP}\|\p - \q\|_2,
\end{equation}
where the first term measures a distance from each output point to the target surface, and the second term intuitively rewards an even distribution on the target surface of the output point cloud \cite{PCN2020}. 
Note that, in Eq.~\eqref{eq:cd_eval}, we use the $\ell 2$ distance, different from the \textit{squared} $\ell 2$ distance used in Eq.~\eqref{eq:cd_train} which is a term in the loss function. 
This is because, computing the $\ell 2$ distance involves square root operation, which is not preferable in the loss function due to its numerical instability.

As our method aims to reconstruct the underlying surface, we also use the point-to-surface distance (P2S): 
\begin{equation}
\mathcal{P}(\tP, \mathcal{S}) = \frac{1}{| \tP |} \sum_{\p \in \tP} \min_{\q \in \mathcal{S}} \| \p - \q \|_2,
\end{equation}
where $\mathcal{S}$ is the underlying surface of the ground truth point cloud $\Pgt$.

These two metrics measure the distance between the denoised point cloud and the ground truth one, with smaller values indicating better results. 

\textbf{Iterative denoising.} For point clouds at higher noise levels, best possible results are obtained by iterative denoising (\ie, feeding the output of the network as the input again), which is similar to previous neural denoising methods such as PCNet and TotalDn. However, compared to them, our method requires much fewer iterations to get the best possible results. We tune the number of iterations for PCNet, TotalDn and our denoiser, and find that for 1\% Gaussian noise, only 1 iteration is required for our denoiser, while 8 iterations are required for PCNet and TotalDn.

\begin{figure*}
  \centering
  \includegraphics[width=\linewidth]{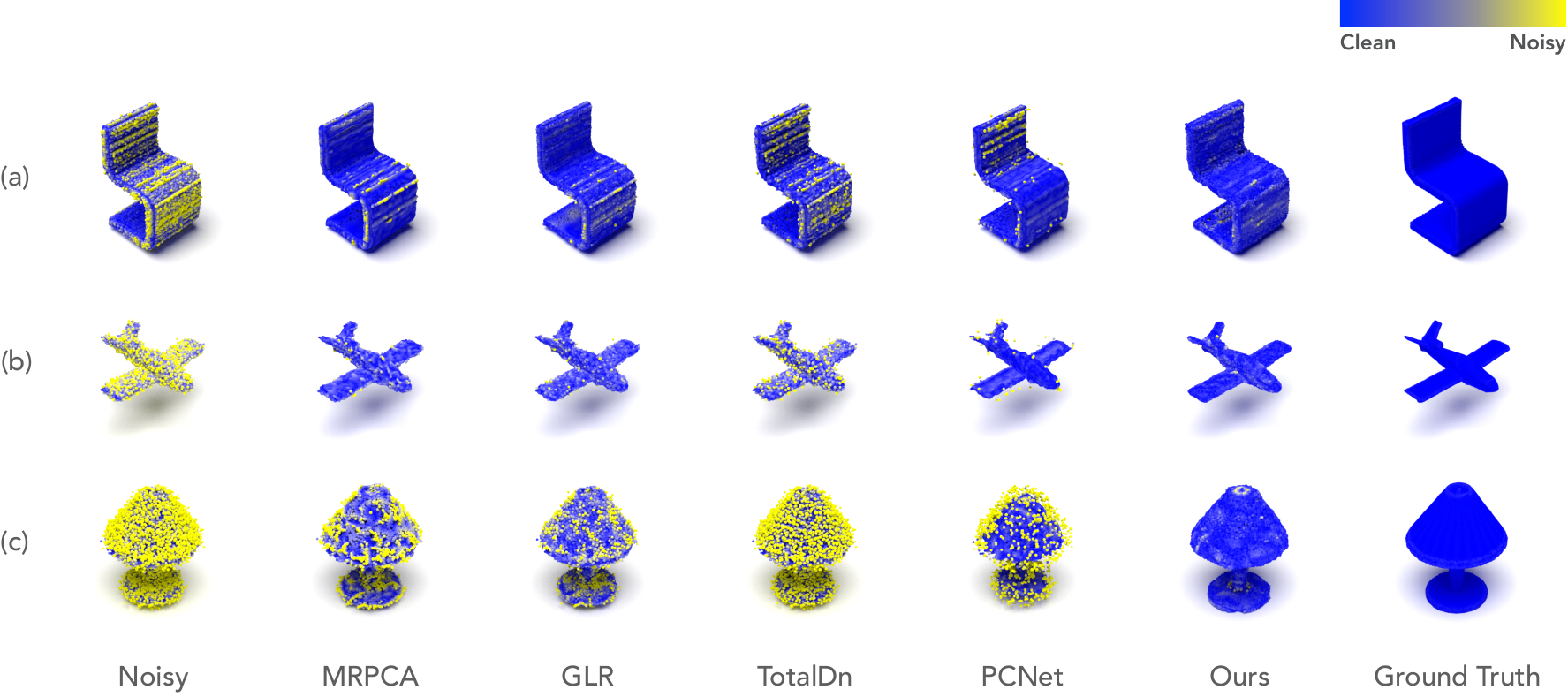}
  \caption{Visual comparison of denoising methods. (a) Simulated scanner noise. (b) 1\% Gaussian noise. (c) 2\% Gaussian noise.}
  \Description{Visual comparison.}
  \label{fig:visual}
\end{figure*}

\subsection{Quantitative Results}
 
We compare both supervised and unsupervised versions of our method quantitatively to state-of-the-art deep-learning based denoising methods as well as non-deep-learning based methods, including PointCleanNet (PCNet) \cite{PCN2020}, TotalDenoising (TotalDn) \cite{TotalDenoising2019}, bilateral filter \cite{bilat}, Jet fitting \cite{jetsfit2005}, MRPCA \cite{MRPCA2017} and GLR \cite{GLR2019}. For each resolution and noise level, we compute the Chamfer distances (CD) and the point-to-surface (P2S) distances based on the 60 point clouds.

Table \ref{tab:comp} shows that the supervised version of our method significantly outperforms previous deep-learning based methods as well as non-deep-learning denoisers. 
The unsupervised version is inferior to our supervised counterpart, but still outperforms Total Denoising, which is also unsupervised, at higher noise levels. 
Also, the unsupervised version performs better than non-deep learning denoisers at 2\%, 2.5\% and 3\% noise levels. 
In general, our method outperforms previous denoising methods, and is more robust to high noise levels.

To examine our method's generalizability, we also conduct evaluations on point clouds perturbed by simulated LiDAR noise. Table \ref{tab:blensor} shows that while our denoiser is trained on Gaussian noise, it is effective in generalizing to the unseen LiDAR noise pattern and performs much better than previous methods.

\textbf{Discussion on results under differnet metrics.}
We notice that the superiority of our method is more significant when measured by the point-to-surface distance (P2S), compared to the Chamfer distance (CD), which is essentially a point-to-point distance. This is because our denoiser reconstructs the underlying manifold of the point cloud and resamples on it. Resampling on the manifold does not guarantee that the newly sampled points are close to the points from the original point cloud, which may lead to comparatively larger point-to-point distances. 
However, the point-to-point distance may not reflect the quality of surface reconstruction well, while the point-to-surface distance generally provides a better measurement as point clouds are representations of 3D surfaces.  

Also, \cite{javaheri2017subjective} finds that the point-to-surface distance is more correlated with subjective evaluation of denoising results. 
The significant superiority in the point-to-surface distance of our method indicates that our method is more visually preferable than previous methods, which is discussed below.

\begin{figure*}
  \centering
  \includegraphics[width=\linewidth]{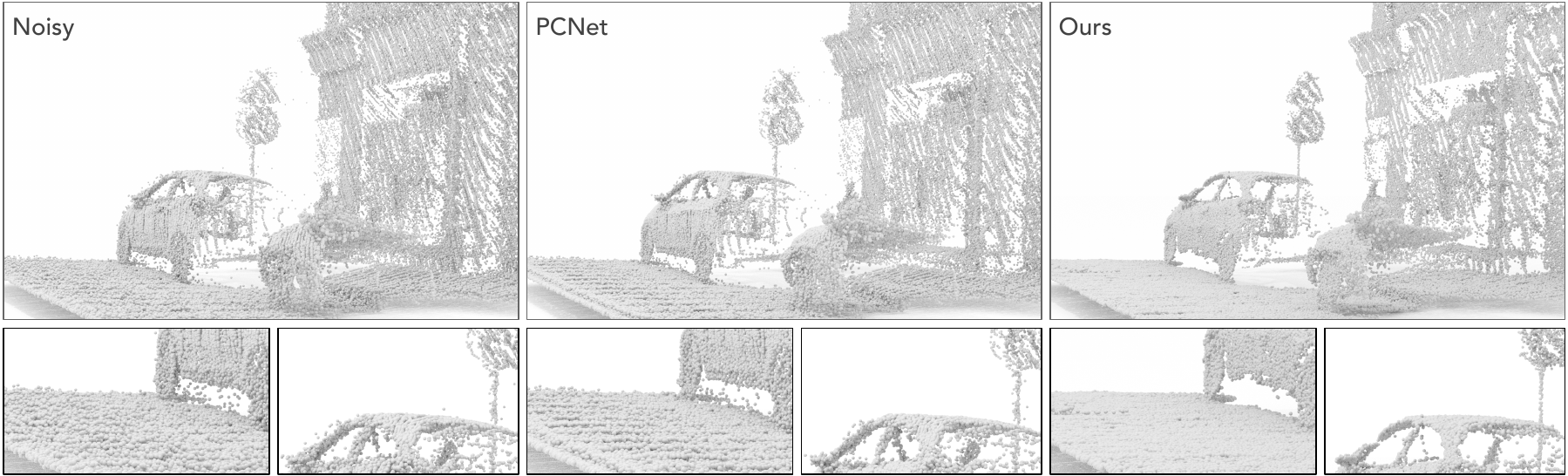}
  \caption{Qualitative results of our denoiser on the real world dataset \textit{Paris-rue-Madame}.}
  \Description{The neural network.}
  \label{fig:rue}
\end{figure*}

\subsection{Qualitative Results}

\begin{figure}
  \centering
  \includegraphics[width=\linewidth]{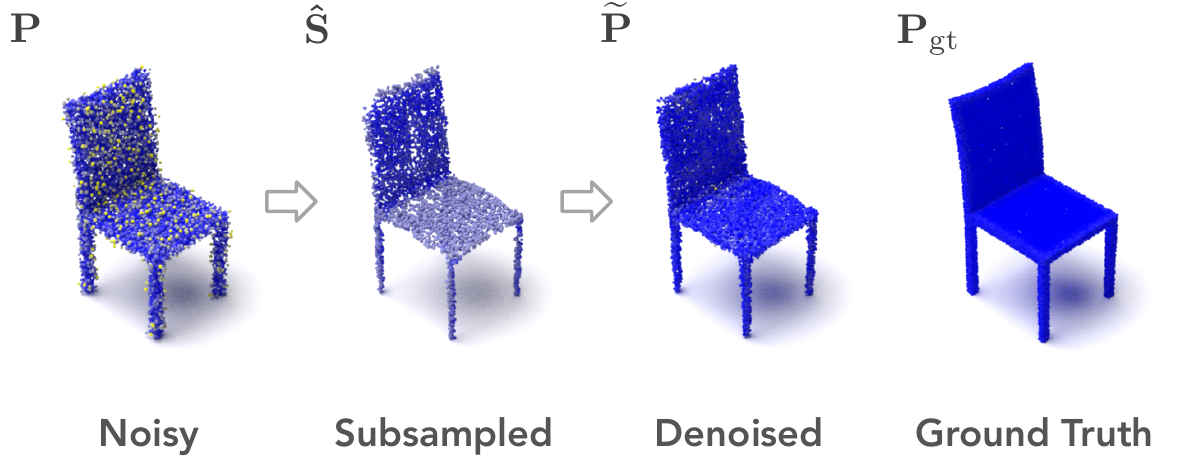}
  \caption{Visualization of the intermediate subsampled point set. Our differentiable pooling operator is effective in sampling points with lower noise.}
  \Description{Visualization of subsampling}
  \label{fig:sample}
\end{figure}

We demonstrate the comparison of visual denoising results under simulated scanner noise and Gaussian noise with different noise levels in Figure \ref{fig:visual}. 
The reconstruction error of each point is measured by the point-to-surface distance. Points with smaller error are colored more blue, and otherwise colored yellow, as indicated in the color bar.
The figure shows that our results are much cleaner and exhibit more visually pleasing surfaces than other methods, especially at higher noise levels. 
Specifically, our method is more robust to outliers compared to the other two deep-learning based methods TotalDn and PCNet. 
Compared to the two state-of-the-art non-deep-learning methods MRPCA and GLR, our method explicitly reconstructs the geometry of the underlying surface and thus can produce results with lower bias. 
In summary, the qualitative results in Figure \ref{fig:visual} are in line with the quantitative results in Table \ref{tab:comp} and \ref{tab:blensor}.

Further, we conduct qualitative studies on the real world dataset {\it Paris-rue-Madame}. 
Note that, the ground truth point cloud is unknown, so the error of each point cannot be visualized as the synthesized datasets.
As demonstrated in Figure \ref{fig:rue}, our denoising result is much cleaner and smoother than that of PCNet, while details are well preserved. 
This validates that our method is effective in generalizing to real world datasets.

In addition, we visualize the intermediate subsampled point set output by the differentiable pooling layer in Fig. \ref{fig:sample}. The figure reveals that our differentiable pooling layer is effective in sampling points with lower noise, which provides a good initialization for the reconstruction of patch manifolds.

\subsection{Ablation Studies}

\begin{table}
  \caption{Ablation studies. All the proposed components contribute positively to the performance.}
  \begin{tabular}{l|ccc|cc}
\toprule
\# & Baseline & Diff. Pool & Dual Loss & 
CD ($10^{-2}$) & P2S ($10^{-2}$) \\
\midrule
1 & \checkmark & & & 
1.41 & 1.72 \\

2 & \checkmark & \checkmark & &
1.36 & 1.63 \\ 

3 & \checkmark & & \checkmark &
1.15 & 1.22 \\ 

\midrule
4 & \checkmark & \checkmark & \checkmark &
\textbf{1.09} & \textbf{1.11} \\
\bottomrule
\end{tabular}
  \label{tab:ablations}
\end{table}

We conduct progressive ablation studies to evaluate the contribution of each component:
\begin{enumerate}
    \item \textbf{Differentiable pooling.} We replace the differentiable pooling layer with a static pooling layer, which downsamples point clouds by random sampling.
    \item \textbf{Dual loss functions.} We remove the Chamfer loss ($\mathcal{L}_\mathrm{CD}$) that explicitly measures the quality of pooling (sampling) and pre-filtering, and employ only the EMD loss ($\mathcal{L}_\mathrm{EMD}$) for final reconstruction. 
\end{enumerate}

The evaluation is based on point clouds of 50K points with 2\% Gaussian noise in our test set. 
As shown in Table \ref{tab:ablations}, all components contribute positively to the full model.

The differentiable pooling enables the denoiser to sample points with lower noise perturbation, relieving the stress of pre-filtering, since the noise level of the input to the pre-filtering layer is lowered. 

The dual loss functions explicitly guide the pre-filtering layer to learn to denoise, leading to a more accurate subset of points that characterizes the underlying manifold, eventually improving the quality of manifold reconstruction.

In summary, the above two components boost the performance of manifold reconstruction, resulting in better denoising output.

\section{Conclusion}

In this paper, we propose a novel paradigm of learning the underlying manifold of a noisy point cloud from differentiably subsampled points. 
We sample points that tend to be closer to the underlying surfaces via an adaptive differentiable pooling operation. 
Then, we infer patch manifolds by transforming each sampled point along with its embedded neighborhood feature to a local surface. 
By sampling on each patch manifold, we reconstruct a clean point cloud that captures the intrinsic structure.   
Our network can be trained end-to-end in either a supervised or unsupervised fashion. 
Extensive experiments demonstrate the superiority of our method compared to the state-of-the-art methods under both synthetic noise and real-world noise.


\bibliographystyle{ACM-Reference-Format}
\bibliography{references}

\appendix

\end{document}